%% file: feffer.tex
\documentclass[letterpaper]{article} 
\usepackage{aaai23}  
\nocopyright
\usepackage{times}  
\usepackage{helvet}  
\usepackage{courier}  
\usepackage[hyphens]{url}  
\usepackage{graphicx} 
\usepackage{natbib}  
\usepackage{caption} 
\usepackage{algorithm}
\usepackage{algorithmic}

\usepackage{newfloat}
\usepackage{listings}
\DeclareCaptionStyle{ruled}{labelfont=normalfont,labelsep=colon,strut=off} 
\floatstyle{ruled}
\newfloat{listing}{tb}{lst}{}
\floatname{listing}{Listing}
\usepackage{microtype}
\usepackage{graphicx}
\usepackage{xcolor}

\usepackage{alltt}
\usepackage{bm}
\usepackage{hhline}
\usepackage{booktabs}
\usepackage{multirow}
\usepackage{subcaption}
\usepackage{makecell}
\usepackage{enumitem}
\setlist{nosep}

\definecolor{background}{HTML}{F7F7F7}
\definecolor{keyword}{HTML}{37AC4A}
\definecolor{operator}{HTML}{A51DFF}
\definecolor{string}{HTML}{C03333}
\definecolor{comment}{HTML}{357979}

\lstdefinelanguage{python}{
  xleftmargin=4mm,
  numbers=left,
  numbersep=2mm,
  numberstyle=\color{gray},
  basicstyle=\fontsize{8}{9}\ttfamily,
  columns=fixed,
  basewidth=0.5em,
  stringstyle=\textcolor{string},
  showstringspaces=false,
  morestring=[b]",
  morestring=[b]""",
  morecomment=[l]\#,
  commentstyle=\textcolor{comment},
  morekeywords={and,as,assert,break,class,continue,def,del,elif,else,except,False,finally,for,from,global,if,import,in,is,lambda,None,nonlocal,not,or,pass,raise,return,True,try,while,with,yield},
  keywordstyle=\color{keyword}\bf\ttfamily,
  literate=
    *{>>}{{\bf\texttt{\color{operator}{>{}>}}}}{1}
    {\&}{{\bf\texttt{\color{operator}{\&}}}}{1}
    {|}{{\bf\texttt{\color{operator}{|}}}}{1}
    {=}{{\bf\texttt{\color{operator}{=}}}}{1}
    {(}{{\bf\texttt{\color{keyword}{(}}}}{1}
    {)}{{\bf\texttt{\color{keyword}{)}}}}{1}
    {[}{{\bf\texttt{\color{keyword}{[}}}}{1}
    {]}{{\bf\texttt{\color{keyword}{]}}}}{1}
    {\{}{{\bf\texttt{\color{keyword}{\char '173}}}}{1}
    {\}}{{\bf\texttt{\color{keyword}{\char '175}}}}{1},
}

\newcommand*{\pyplain}[1]{\fontsize{8}{9}\texttt{#1}}
\newcommand*{\scriptsf}[1]{\textsf{\scriptsize #1}}

\newcommand{\stdevfont}{\fontsize{8}{8}\selectfont\par}

\definecolor{darkgreen}{HTML}{228B22}
\newcommand{\vclosetohopt}[1]{\textcolor{darkgreen}{#1}}
\newcommand{\closetohopt}[1]{\textcolor{blue}{#1}}

\usepackage{amsmath}
\usepackage{amssymb}
\usepackage{mathtools}
\usepackage{amsthm}

\usepackage[capitalize,noabbrev]{cleveref}

\theoremstyle{plain}

\theoremstyle{definition}

\theoremstyle{remark}

\usepackage[textsize=tiny]{todonotes}

\title{Navigating Ensemble Configurations for Algorithmic Fairness}
\author {
    Michael Feffer,\textsuperscript{\rm 1}
    Martin Hirzel,\textsuperscript{\rm 2}
    Samuel C.~Hoffman,\textsuperscript{\rm 2}\\
    Kiran Kate,\textsuperscript{\rm 2}
    Parikshit Ram,\textsuperscript{\rm 2}
    Avraham Shinnar\textsuperscript{\rm 2}
}
\affiliations {
    \textsuperscript{\rm 1} Carnegie Mellon University, Pittsburgh, PA, USA\\
    \textsuperscript{\rm 2} IBM Research, Yorktown Heights, NY, USA\\
    mfeffer@andrew.cmu.edu, hirzel@us.ibm.com
}

\begin{document}

\maketitle

\begin{abstract}
\input{abstract}
\end{abstract}


\input{intro}
\input{related}
\input{library}
\input{methodology}
\input{results}
\input{discussion}
\input{conclusion}

\bibliography{bibfile}

\clearpage
\input{appendix}

\end{document}

%% file: abstract.tex
Bias mitigators can improve algorithmic fairness in
machine learning models, but their effect on fairness
is often not stable across data splits.
A popular approach to train more stable models is ensemble learning,
but unfortunately, it is unclear how to combine ensembles with mitigators
to best navigate trade-offs between fairness and predictive performance.
To that end, we built an open-source library enabling the modular composition of
8~mitigators, 4~ensembles, and their corresponding hyperparameters,
and we empirically explored the space of configurations on 13 datasets.
We distilled our insights from this exploration in the form
of a guidance diagram for practitioners that we demonstrate is
robust and reproducible.


%% file: intro.tex
\section{Introduction}\label{sec:intro}

Algorithmic bias in machine learning can lead to models that discriminate
against underprivileged groups in various domains, including hiring,
healthcare, finance, criminal justice,
education, and even child care.
Of course, bias in machine learning is a socio-technical problem that
cannot be solved with technical solutions alone.
That said, to make tangible progress, this paper focuses on
\emph{bias mitigators}, which improve or replace an existing
machine learning estimator (e.g., a classifier) so it makes less
biased predictions (e.g.,~class labels)
as measured by a fairness metric (e.g., disparate impact~\cite{feldman_et_al_2015}).
%
Unfortunately, bias mitigation often suffers from high \emph{volatility},
meaning
the estimator is less stable
with respect to group fairness metrics. 
In the worst case, this volatility
can even cause a model to appear fair when measured on training data
while being unfair on production data.
Given that ensembles (e.g., bagging or boosting) can improve stability
for accuracy metrics~\cite{witten_et_al_2016},
we felt it was important to explore whether they also
improve stability for group fairness metrics.

%
%

Unfortunately, the sheer number of ways in which ensembles and
mitigators can be combined and configured with base estimators and
hyperparameters presents a dilemma.
On the one hand, the diversity of the space increases the chances of
it containing at least one combination with satisfactory fairness
and/or predictive performance for the provided data.
On the other hand, finding this combination via brute-force
exploration may be untenable if resources are limited.

To this end,
we conducted experiments that navigated this space with
8~bias mitigators from AIF360~\cite{bellamy_et_al_2018};
bagging, boosting, voting, and stacking ensembles from the popular
scikit-learn library~\cite{buitinck_et_al_2013};
and 13~datasets of various sizes and baseline fairness
(earlier papers used at most a handful).
Specifically, we searched the Cartesian product of 
datasets, mitigators, ensembles, and hyperparameters both via
brute-force and via Hyperopt \cite{bergstra2013hyperopt}
for configurations that optimized fairness while maintaining
decent predictive performance and vice-versa.
Our findings confirm the intuition that ensembles often improve
stability of not just accuracy but also of the group fairness metrics
we explored.
However, the best configuration of mitigator and ensemble depends on
dataset characteristics, evaluation metric of choice, and even
worldview~\cite{friedler_scheidegger_venkatasubramanian_2021}.
Therefore, we automatically distilled a method selection guidance diagram
in accordance with the results from both brute-force search
and Hyperopt exploration.

To support these experiments, we assembled a library of pluggable
ensembles, bias mitigators, and fairness datasets.
While we reused popular and well-established
open-source technologies, 
we made several new adaptations in our library to get components to work
well together.
Our library is available open-source
(\url{https://github.com/IBM/lale})
to encourage research and
real-world adoption.
%
%



%% file: related.tex
\section{Related Work}\label{sec:related}

A few pieces of prior work used ensembles for
fairness, but they used specialized ensembles and bias mitigators,
in contrast to our work, which uses off-the-shelf modular components.
The \emph{discrimination-aware ensemble} uses a heterogeneous
collection of base estimators~\cite{kamiran_karim_zhang_2012};
when they all agree, it returns the consensus
prediction, otherwise, it classifies instances as positive iff
they belong to the unprivileged group. 
%
The \emph{random ensemble} also uses a heterogeneous collection of
base estimators, and picks one of them at random to make a
prediction~\cite{grgichlaca_et_al_2017}. 
The paper offers a
synthetic case where the ensemble is more fair and more
accurate than all base estimators, but lacks experiments with real
datasets.
\emph{Exponentiated gradient reduction} trains a sequence of base
estimators using a game theoretic model where one player seeks to maximize fairness
violations by the estimators so far and the other player seeks to
build a fairer next estimator~\cite{agarwal_et_al_2018}.  In the end,
for predictions, it uses weights to pick a random base estimator.
%
\emph{Fair AdaBoost} modifies boosting to
boost not for accuracy but for individual fairness~\cite{bhaskaruni_hu_lan_2019}. 
In the end, for predictions, it gives a base estimator
higher weight if it was fair on more instances from the
training set.
The \emph{fair voting ensemble} uses a heterogeneous collection of
base estimators~\cite{kenfack_et_al_2021}. Each prediction
votes among base estimators $\phi_t$, \mbox{$t\in 1..n$}, with weights
\mbox{$W_t=\alpha\cdot A_t/(\Sigma_{t=1}^nA_j)+(1-\alpha)\cdot F_t/(\Sigma_{t=1}^nF_j)$},
where $A_t$ is an accuracy metric and $F_t$ is a fairness metric.
The \emph{fair double ensemble} uses stacked predictors,
where the final estimator is linear, with a novel approach to train
the weights of the final estimator to satisfy a system of accuracy and
fairness constraints~\cite{mishler_kennedy_2021}.

Each of the above-listed approaches used an ensemble-specific bias mitigator,
whereas we experiment with eight different off-the-shelf modular
mitigators.  Moreover, each of these approaches used one specific
kind of ensemble, whereas we experiment with off-the-shelf modular
implementations of bagging, boosting, voting, and stacking.
Using off-the-shelf mitigators and ensembles facilitates plug-and-play
between the best available independently-developed implementations.
Out of the work on fairness with ensembles discussed above,
one paper had an experimental
evaluation with five datasets~\cite{agarwal_et_al_2018} and the other
papers used at most three datasets. In contrast, we use
13~datasets. Finally, unlike these earlier papers, our paper
specifically explores fairness stability and the best ways
to combine mitigators and ensembles. We auto-generate a guidance diagram
from this exploration.

Ours is not the first paper to use automated machine learning,
including Bayesian optimizers, to optimize models and mitigators for
fairness~\cite{perrone2020bayesian,wu2021fair}.
But unlike prior work, we specifically focus on
applying AutoML to ensemble learning and bias mitigation to
validate our guidance diagram and other search.

Our work takes inspiration from earlier empirical studies of fairness
techniques
\cite{biswas_rajan_2021,friedler_et_al_2019,holstein_et_al_2019,lee_singh_2021,singh_et_al_2021,valentim_lourencco_antunes_2019,yang_et_al_2020},
which help practitioners and researchers better understand the state
of the art. But unlike these works, we experiment with ensembles and
with fairness stability.

Our work also offers a new library of bias mitigators.
While there have been excellent prior fairness toolkits such as
ThemisML~\cite{bantilan_2017}, AIF360~\cite{bellamy_et_al_2018}, and
FairLearn~\cite{agarwal_et_al_2018}, none of them support ensembles.
Ours is the first that is modular enough to investigate a large space of
unexplored mitigator-ensemble combinations.
%
We previously published some aspects of our library in a non-archival
workshop with no official proceedings, but that paper did not yet
discuss ensembles~\cite{hirzel_kate_ram_2021}.
In another non-archival workshop paper, we discussed ensembles and
some of these experimental results \cite{feffer_et_al_2022}, but no
Hyperopt results and only limited analysis of
the guidance diagram, both of which are present in this work.


%% file: library.tex
\input{stdmodel}

\section{Library and Datasets}\label{sec:library}



Aside from our experiments, one contribution of our work is
implementing compatibility between mitigators from
AIF360~\cite{bellamy_et_al_2018} and ensembles from
scikit-learn~\cite{buitinck_et_al_2013}.
To provide the glue and facilitate searching over a space of mitigator
and ensemble configurations, we extended the Lale open-source library
for semi-automated data science~\cite{baudart_et_al_2021}.

\paragraph{Metrics.}
This paper uses metrics from scikit-learn, including precision, recall,
and $F_1$ score. 
In addition, we implemented a scikit-learn compatible API for
several fairness metrics from AIF360 including
disparate impact (as described
in~\citet{feldman_et_al_2015}).
We also measure time (in seconds) and memory (in MB)
utilized when fitting models.




\paragraph{Ensembles.}
Ensemble learning uses multiple weak models to form one strong model.
%
Our experiments use four ensembles supported by scikit-learn: 
bagging, boosting, voting, and stacking.
Following scikit-learn, we use the following terminology to characterize ensembles:
A \emph{base estimator} is an estimator that serves as a building
block for the ensemble.
An ensemble supports one of two \emph{composition} types: whether the
ensemble consists of identical base estimators (\emph{homogeneous}, e.g.\ bagging and boosting) or
different ones (\emph{heterogeneous}, e.g.\ voting and stacking).
%
%
For the homogeneous ensembles,
we
used their most common base estimator in practice: the
decision-tree classifier.
For the heterogeneous ensembles (voting and stacking), we used a set
of typical base estimators: XGBoost~\cite{chen_guestrin_2016}, random
forest, k-nearest neighbors, and support vector machines.
Finally, for stacking, we also used XGBoost as the final estimator.

\paragraph{Mitigators.}

We added support in Lale for bias mitigation from AIF360~\cite{bellamy_et_al_2018}.
AIF360 distinguishes three kinds of mitigators for improving group fairness:
\emph{pre-estimator mitigators}, which are learned input manipulations
that reduce bias in the data sent to downstream estimators
(we used 
DisparateImpactRemover~\cite{feldman_et_al_2015},
LFR~\cite{zemel_et_al_2013},
and Reweighing~\cite{kamiran_calders_2012});
\emph{in-estimator mitigators}, which are specialized estimators that
directly incorporate debiasing into their training
(AdversarialDebiasing~\cite{zhang_lemoine_mitchell_2018},
GerryFairClassifier~\cite{kearns_et_al_2018},
MetaFairClassifier~\cite{celis_et_al_2019},
and PrejudiceRemover~\cite{kamishima_et_al_2012});
and
\emph{post-estimator mitigators}, which reduce bias in
predictions made by an upstream estimator
(we used CalibratedEqOddsPostprocessing~\cite{pleiss_et_al_2017}).

Fig.~\ref{fig:standard_model} visualizes the combinations of
ensemble and mitigator kinds we explored,
while also highlighting the modularity of our approach.
Mitigation strategies can be applied at the level of either the base
estimator or the entire ensemble, but by the nature of some ensembles
and mitigators, not all combinations are feasible.
%
First, post-estimator mitigators typically do not support \pyplain{predict\_proba} functionality required
for some ensemble methods and recommended for others. Calibrating probabilities
from post-estimator mitigators has been
shown to be tricky~\cite{pleiss_et_al_2017}, so despite Lale support
for other post-estimator mitigators, our experiments only explored
CalibratedEqOddsPostprocessing.
Additionally, it is impossible to apply an in-estimator mitigator at the ensemble level,
so we exclude those combinations.
Finally, we decided to omit some combinations that are technically
feasible but less interesting.  For example, while our library
supports mitigation at multiple points, say, at both the ensemble
and estimator level of bagging, we elided these configuration from
Fig.~\ref{fig:standard_model} and from our experiments.

\input{datasets}

\paragraph{Datasets.}

We gathered the datasets for our experiments primarily from OpenML~\cite{vanschoren_et_al_2014};
the exceptions come from Medical Expenditures Panel Survey (MEPS) data 
not hosted there \cite{meps2015,meps2016}. Some
have been used extensively as benchmarks elsewhere in the algorithmic fairness literature. We pulled other novel datasets 
from OpenML that have demographic data that
could be considered protected attributes (such as race, age, or gender) and contained associated baseline levels of disparate impact.
In addition, to get a sense for the predictive power of each protected attribute,
we fit XGBoost models to each dataset with five different seeds and found the ranking of the 
average feature importance (where 1 is the most important) of the most predictive protected attribute
for that dataset.
In all, we used 13 datasets, with most information 
summarized in Table~\ref{tab:dataset-table}
and granular feature importance information summarized in the Appendix.
When running experiments,
we split the datasets using stratification by not just the target
labels but also the protected attributes~\cite{hirzel_kate_ram_2021},
leading to moderately more homogeneous
fairness results across different splits.
The exact details of the preprocessing
are in the open-source code for our library for reproducibility.
We hope that bundling these datasets and default preprocessing
with our package, in addition to AIF360 and scikit-learn compatibility,
will improve dataset quality going forward. 




%% file: stdmodel.tex
\begin{figure*}[!t]
  \centerline{\includegraphics[width=0.9\textwidth]{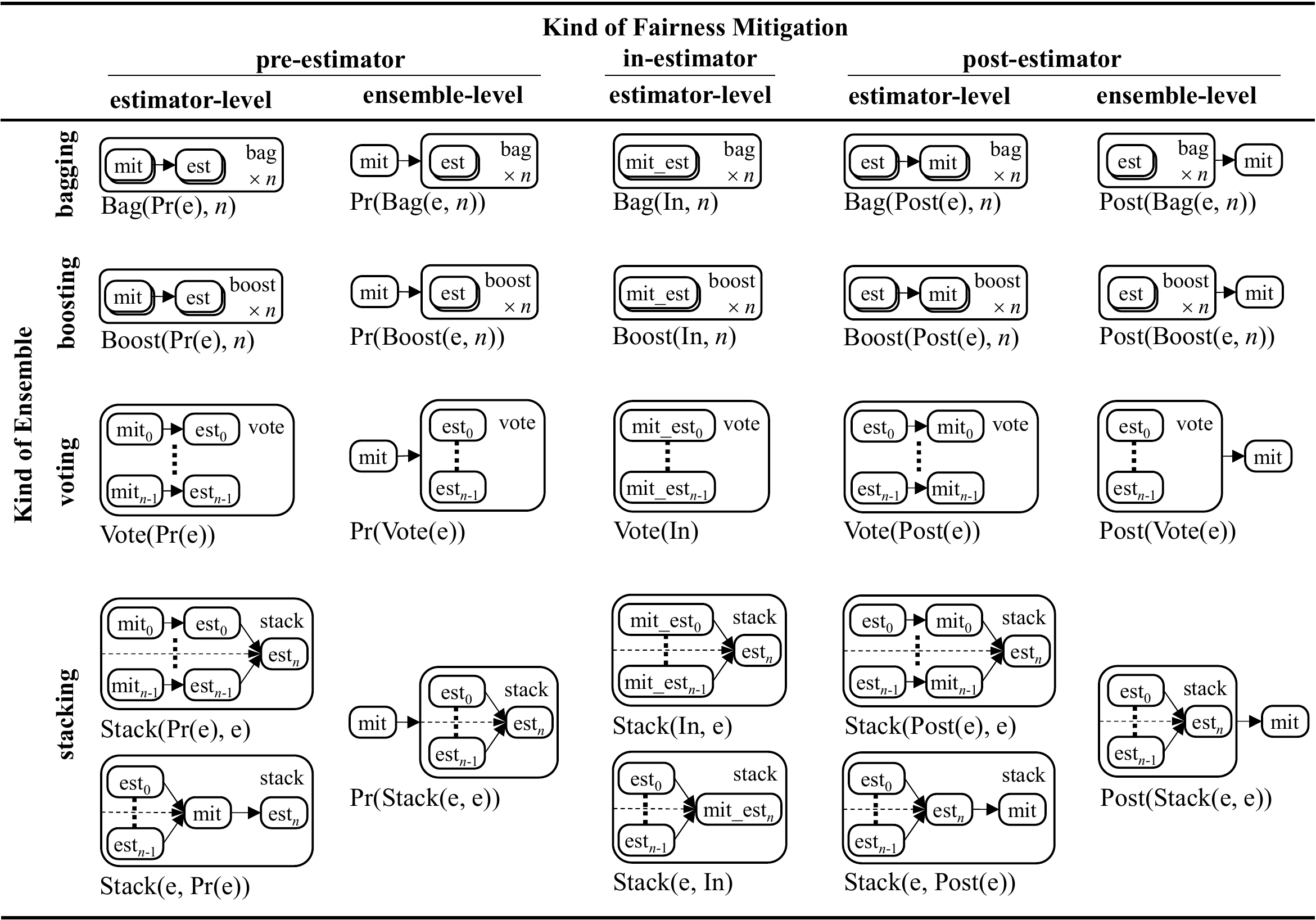}}
  \caption{\label{fig:standard_model}Combinations of ensembles and mitigators.
  \scriptsf{Pr(e)} applies a pre-estimator mitigator before an
estimator~\scriptsf{e};
\scriptsf{In} denotes an in-estimator mitigator, which is itself an
estimator;
and \scriptsf{Post(e)} applies a post-estimator mitigator after an
estimator~\scriptsf{est}.
\mbox{\scriptsf{Bag(e, n)}} is short for \scriptsf{BaggingClassifier} with
\scriptsf{n} instances of base estimator \scriptsf{e};
\scriptsf{Boost(e, n)} is short for \scriptsf{AdaBoostClassifier} with
\scriptsf{n} instances of base estimator~\scriptsf{e};
\scriptsf{Vote(e)} applies a \scriptsf{VotingClassifier} to a list
of base estimators \scriptsf{e};
and \scriptsf{Stack(e, e)} applies a
\scriptsf{StackingClassifier} to a list of base estimators
(first~\scriptsf{e}) and a final estimator (second~\scriptsf{e}).
    For stacking, the \scriptsf{passthrough} option is represented by a
    dashed arrow.}
\end{figure*}

%% file: datasets.tex
\begin{table*}[!ht]
  \centerline{\small\begin{tabular}{lllrrrr}
  \toprule
  \textbf{Dataset} & \textbf{Description} & \textbf{Privileged group(s)} & \textbf{Importance} & $\bm{N_\textit{cols}}$ & $\bm{N_\textit{rows}}$ & \textbf{DI}\\
  \midrule
  COMPAS Violent   & Correctional offender violent recidivism     & White women                  &    4 &    10 &    3,377 &   0.822\\
  Credit-g         & German bank data quantifying credit risk     & Men and older people         &   22 &    58 &    1,000 &   0.748\\
  COMPAS           & Correctional offender recidivism             & White women                  &    5 &    10 &    5,278 &   0.687\\
  Ricci            & Fire department promotion exam results       & White men                    &    6 &     6 &      118 &   0.498\\
  TAE              & University teaching assistant evaluation     & Native English speakers      &    1 &     6 &      151 &   0.449\\
  Titanic          & Survivorship of Titanic passengers           & Women and children           &    2 &    37 &    1,309 &   0.263\\
  \midrule                                                                                                                    
  SpeedDating      & Speed dating experiment at business school   & Same race                    &   24 &    70 &    8,378 &   0.853\\
  Bank             & Portuguese bank subscription predictions     & Older people                 &   17 &    51 &   45,211 &   0.840\\
  MEPS 19          & Utilization results from Panel 19 of MEPS    & White individuals            &   22 &   138 &   15,830 &   0.490\\
  MEPS 20          & Same as MEPS 19 except for Panel 20          & White individuals            &   18 &   138 &   17,570 &   0.488\\
  Nursery          & Slovenian nursery school application results & ``Pretentious parents''      &    3 &    25 &   12,960 &   0.461\\
  MEPS 21          & Same as MEPS 19 except for Panel 21          & White individuals            &   10 &   138 &   15,675 &   0.451\\
  Adult            & 1994 US Census salary data                   & White men                    &   19 &   100 &   48,842 &   0.277\\
  \bottomrule
  \end{tabular}}
  \caption{Qualitative and quantitative summary information of the datasets. The datasets are ordered by first partitioning by whether they contain at least 8,000 rows
  (we picked 8,000 to get a roughly even split; the partition is represented by the horizontal line in the middle of the table) and then sorting by descending baseline disparate impact (DI).
  Values for feature importance ranking of most predictive protected attribute according to XGBoost (Importance), the number of columns ($N_\textit{cols}$), number of rows ($N_\textit{rows}$), and baseline disparate impact (DI) displayed here are computed \emph{after} preprocessing techniques are applied.}
  \label{tab:dataset-table}
\end{table*}

%% file: methodology.tex
\section{Methodology}\label{sec:methodology}

Given our 13~datasets, 4~types of ensembles, 8~mitigators, and
all relevant hyperparameters, we wanted to gain insights about
the best ways to combine ensemble learning and bias mitigation
in various problem contexts and data setups. To this end,
we conducted two searches over the Cartesian product of
these settings and compared their results.
The first was a manual grid search to determine
optimal configurations for each dataset. The second
also involved finding suitable configurations per dataset,
but was
automated via Bayesian optimization in Hyperopt~\cite{bergstra2013hyperopt}.


\subsection{Grid Search}






We organize our grid search experiments into two steps.
The first is a preliminary search that finds ``best'' mitigators
without ensembles.
The second is the ensemble experiments using the mitigator configurations
selected by the first.


\paragraph{First step.}
It is difficult to define ``best'' (in an
empirical sense) given
different dimensions of performance and datasets.
To this end, we first run grid searches over each dataset,
exploring mitigators and their hyperparameters with
basic decision-trees where needed.
We run 5~trials of 3-fold cross validation for each configuration.
For each dataset, we choose a ``best'' pre-, \mbox{in-,} and post-estimator mitigator:


\begin{enumerate}
  \item Filter configurations to ones with acceptable fairness,
    defined as mean disparate impact between 0.8 and 1.25.
  \item Filter remaining to ones with nontrivial precision.
  \item Filter remaining to ones with good
    predictive performance, defined as mean $F_1$ score (across 5 trials)
    greater than the average of all mean $F_1$ scores
    or the median of all mean $F_1$ scores, whichever is greater.
  \item Finally, select the mitigator with maximum
    precision (in case of COMPAS, since true positives should be
    prioritized) or recall (all other datasets, since false negatives
    should be avoided).
\end{enumerate}

\noindent
Tables~\ref{tab:pre-est-configs} and~\ref{tab:in-est-configs} in our
Appendix list the chosen pre-estimator and in-estimator configurations
(the only post-estimator configuration is
CalibratedEqOddsPostprocessing).


\paragraph{Second step.}
Given the ``best'' mitigator configurations,
this step explores the Cartesian product of ensembles and
mitigators of
Fig.~\ref{fig:standard_model} plus ensemble hyperparameters.
For bagging and boosting,
the only ensemble-level hyperparameter varied between configurations
was the number of base estimators:
$\{1,10,100\}$ for bagging and $\{1,50,500\}$ for boosting.
Voting and stacking use lists of heterogeneous base estimators as hyperparameters.
In our experiments, these lists contained either $4$ mitigated or $4$ unmitigated
base estimators.
For the in-estimator mitigation case
these were \{PrejudiceRemover, GerryFairClassifier, MetaFairClassifier, and AdversarialDebiasing\}.
Lastly, stacking also has a \pyplain{passthrough} hyperparameter controlling whether
dataset features were passed to the final estimator.
If \pyplain{passthrough} is set to \pyplain{False}, it is impossible to mitigate the final estimator
due to lack of dataset features; otherwise
we mitigate either the base estimators
or final estimator, but not both.
The second step also uses 5~trials of 3-fold
cross validation for each experiment, running
%
on a computing cluster with Intel Xeon E5-2667 processors @ 3.30GHz.
Every experiment configuration run was allotted 4 cores and 12 GB memory.

\subsection{Hyperopt Search}

\begin{figure}[!b]
\begin{lstlisting}[language=python]
def symm_di(model, X, y):  # symmetric disparate impact
    di = di_scorer(model, X, y)
    return di if di <= 1 else 1 / di
min_di, max_di = symm_di.score_data(X=X, y_pred=y), 1
min_f1 = f1_scorer(dummy, X, y) # constant positive label
max_f1 = f1_scorer(xgboost, X, y) # use gradient boosting
def blended_scorer(model, X, y):
    di, f1 = symm_di(model, X, y), f1_scorer(model, X, y)
    di = (di - min_di) / (max_di - min_di)  # scale DI
    f1 = (f1 - min_f1) / (max_f1 - min_f1)  # scale F1
    if di < 0.66: di -= 0.66 - di  # amplify DI if low
    if f1 < 0.66: f1 -= 0.66 - f1  # amplify F1 if low
    return 0.5 * (di + f1)  # blend to joint objective
\end{lstlisting}
\caption{\label{fig:blended_scorer}Blended objective for Hyperopt search.}
\end{figure}

We used Hyperopt to perform another model configuration search, this
time in a single step guided by an objective that combined predictive
performance and fairness.
We defined a single search space that includes all ensembles and
mitigators and their hyperparameters.
Then, we defined the blended scorer in Fig.~\ref{fig:blended_scorer}.
Finally, we ran Lale's Hyperopt wrapper, passing the
\pyplain{blended\_scorer} as the objective to maximize and setting
timeouts of 10~minutes per trial and 20~hours total for each dataset,
on the same cluster as for grid search.


%% file: results.tex
\section{Results}\label{sec:results}

This section includes quantitative results of our two searches
and qualitative guidance regarding future model development
based on these results.

\subsection{Grid Search Results}

\paragraph{Result preprocessing.}

To facilitate cross-dataset comparisons, we applied the following
procedure on a per-dataset basis for each metric of interest: 
(i)~given all results, map all values to the same region of metric space around the point of optimal fairness
(i.e. for disparate impact, we use the reciprocal of a value if it is larger than 1 for downstream calculations,
and for statistical parity difference, we use the absolute value),
and (ii)~min-max scale the mean and standard deviation
of the metric of interest, separately. After doing this for all datasets, we
group remaining results by mitigator kind and ensemble type, and average the scaled values
over all datasets for each group. 
Given a metric~$M$, we refer to the result of this procedure
using mean values as ``standardized $M$
outcome'' and using standard deviation as
``standardized $M$ volatility''.
The tables and figures that follow report values normalized as described above.

\paragraph{Do ensembles help with fairness?}

Table~\ref{tab:rq1} shows the disparate
impact results. Mitigation
almost always improved disparate impact outcomes,
but ensemble learning generally incurred a slight penalty relative
to the no-ensemble baseline.
However, ensemble learning does generally reduce disparate impact volatility.
In some contexts,
this increased stability may be preferred over
better yet more unstable predictions.

\begin{table}[]
  \setlength{\tabcolsep}{2.6pt}
  \begin{tabular}{@{}lcccccccc@{}}

  \toprule
              & \multicolumn{2}{c}{No Mit.}    & \multicolumn{2}{c}{Pre-}        & \multicolumn{2}{c}{In-}         & \multicolumn{2}{c}{Post-}       \\
              \cmidrule(rl){2-3}                \cmidrule(rl){4-5}                \cmidrule(rl){6-7}                \cmidrule(rl){8-9}
              & DO            & DV            & DO            & DV            & DO            & DV            & DO            & DV            \\
  \midrule
  No ensemble & \textbf{0.42} & 0.18          & 0.73          & 0.38          & \textbf{0.87} & 0.44          & \textbf{0.53} & 0.24          \\
  \addlinespace[0.5em]
  Bagging     & 0.31          & \textbf{0.08} & 0.54          & \textbf{0.19} & 0.80          & 0.28          & 0.44          & \textbf{0.08} \\
  Boosting    & 0.33          & 0.18          & \textbf{0.69} & 0.39          & \textbf{0.87} & \textbf{0.26} & 0.41          & 0.12          \\
  Voting      & 0.29          & 0.09          & 0.51          & 0.35          & 0.40          & 0.45          & 0.21          & 0.20          \\
  Stacking    & 0.39          & 0.19          & 0.61          & 0.27          & 0.44          & 0.39          & 0.50          & 0.27          \\
  \bottomrule
  \end{tabular}
  \caption{Standardized Disparate impact Outcome (DO) and Volatility (DV).
  Note that DO and DV utilize different scales.}
  \label{tab:rq1}
  \end{table}

\paragraph{Do ensembles help predictive performance when there is mitigation?}

Table~\ref{tab:rq2} shows $F_1$ results. Even with ensemble learning, mitigation decreases predictive performance,
but relative to standalone mitigators, mitigated ensembles typically have better outcomes or
stability, but not both.
Except for a few cases, mitigated ensembles \emph{can} help with
predictive performance \emph{or} $F_1$ volatility.

\begin{table}[]
  \setlength{\tabcolsep}{2.6pt}
  \begin{tabular}{@{}lcccccccc@{}}
  \toprule
              & \multicolumn{2}{c}{No Mit.}     & \multicolumn{2}{c}{Pre-}        & \multicolumn{2}{c}{In-}         & \multicolumn{2}{c}{Post-}       \\
              \cmidrule(rl){2-3}                \cmidrule(rl){4-5}                \cmidrule(rl){6-7}                \cmidrule(rl){8-9}
              & FO            & FV            & FO            & FV            & FO            & FV            & FO            & FV            \\
  \midrule
  No ensemble & 0.70          & 0.20          & 0.54          & 0.39          & 0.51          & 0.49          & 0.63          & 0.19          \\
  \addlinespace[0.5em]
  Bagging     & \textbf{0.93} & 0.13          & 0.50          & \textbf{0.19} & 0.61          & \textbf{0.11} & 0.65          & \textbf{0.13} \\
  Boosting    & 0.84          & 0.28          & 0.49          & 0.25          & 0.52          & 0.28          & 0.63          & \textbf{0.13} \\
  Voting      & 0.77          & \textbf{0.09} & 0.40          & 0.36          & 0.45          & 0.50          & 0.58          & 0.19          \\
  Stacking    & 0.83          & 0.26          & \textbf{0.56} & 0.50          & \textbf{0.67} & 0.59          & \textbf{0.66} & 0.27          \\
  \bottomrule
  \end{tabular}
  \caption{Standardized $F_1$ outcome (FO) and volatility (FV).}
  \label{tab:rq2}
\end{table}

\input{fig_tree}

\subsection{Guidance for method selection}


To advise future practitioners based on our results, we generated
Fig.~\ref{fig:tree} from optimal configurations for particular metrics and data setups.
To generate it, we do the following:



\begin{enumerate}
    \item Organize all results by dataset. 
    \item Filter results for each dataset to ones that occur in the top 33\% of results for both standardized disparate impact outcome and standardized $F_1$ outcome.
    \item Place each result into one of four quadrants based on the dataset's baseline fairness and size.
    \item Average each metric in each quadrant while grouping by model configuration.
    \item Report the top~3 configurations per quadrant and metric.
\end{enumerate}






\paragraph{Leave-one-out evaluation.}

One way in which we evaluate our guidance diagram is, for each
dataset, to follow the 
diagram generation steps while leaving out the results pertaining to
that dataset, and examine differences in terms of the recommended
model configurations and their performances between the new diagram and the
one generated from all of the datasets. Because our guidance
diagram has three recommendations per metric, the largest number of differences
between a leave-one-out diagram and the full dataset diagram for a given metric
is three. We also compute signed differences of metric values by subtracting
the metric value of the best model recommended by the leave-one-out diagram
from that of the full dataset diagram.
However, if the diagram creation method generalizes well
and the diagram itself has reproducible insights, these differences should be
close to zero.
Table \ref{tab:leave-one-out-results} displays both types of these differences
for all omitted datasets and the metrics disparate impact mean, disparate impact standard deviation,
and $F_1$ mean. Based on these differences, some datasets
have more of an effect on the guidance diagram than others. This phenomenon
will be covered in our discussion section.

\begin{table}
    \setlength{\tabcolsep}{1.9pt}
    \begin{tabular}{lcccccc}
    \toprule
                    & \multicolumn{2}{c}{DI Mean} & \multicolumn{2}{c}{DI StdDev} & \multicolumn{2}{c}{F1 Mean} \\
                    \cmidrule(rl){2-3} \cmidrule(rl){4-5} \cmidrule(rl){6-7}
    Omitted Dataset & Num & Metric & Num & Metric & Num & Metric \\
    \midrule
    COMPAS Violent & 0       & 0            & 0         & 0              & 0       & 0             \\ 
    Credit-g       & 3       & 0.29         & 3         & -0.03       & 3       & 0.23      \\ 
    COMPAS          & 0       & 0            & 0         & 0              & 0       & 0             \\ 
    Ricci           & 0       & 0            & 0         & 0              & 0       & 0             \\ 
    TAE             & 2       & 0.20         & 1         & 0              & 1       & -0.19      \\ 
    Titanic         & 1       & 0            & 1         & -0.12       & 1       & 0             \\ 
    SpeedDating     & 0       & 0            & 2         & 0              & 3       & 0.05      \\ 
    Bank            & 3       & 0.11         & 1         & -0.01       & 1       & 0             \\ 
    MEPS 19        & 0       & 0            & 0         & 0              & 0       & 0             \\ 
    MEPS 20        & 1       & 0.01         & 2         & 0       & 0       & 0      \\ 
    Nursery         & 0       & 0            & 0         & 0              & 0       & 0             \\ 
    MEPS 21        & 1       & -0.04        & 1         & 0.03       & 1       & 0      \\ 
    Adult           & 0       & 0            & 0         & 0              & 0       & 0             \\
    \bottomrule
    \end{tabular}
    \caption{Number of configuration differences and signed metric differences between 
    leave-one-out and full dataset guidance diagrams for each omitted dataset.
    Note that metric differences are \textit{not} standardized values.}
    \label{tab:leave-one-out-results}
\end{table}

\subsection{Hyperopt Result Comparison}

We purposely designed a scoring function for Hyperopt (see methodology
section) that is is similar to the method we used to filter grid
search results to produce the guidance diagram. Therefore, Hyperopt's solutions
provide another way to evaluate the guidance diagram's suggested configurations.

Table~\ref{tab:hyperopt-and-guidance-recs} shows the configurations returned by Hyperopt
for each dataset as well as corresponding average disparate impact and F1 score
with standard deviations. For comparison, it also shows the same values for the configurations
recommended by the guidance diagram when average disparate impact or 
average F1 score is the metric of interest. A close inspection of
the table
reveals that while the guidance diagram rarely recommends
the exact same configuration as that found by Hyperopt, it can recommend
one with approximately the same performance.

\begin{table*}[!htbp]
    \setlength{\tabcolsep}{3.2pt}
    \fontsize{9}{9}\selectfont\par
    \begin{tabular}{lccccccccc}
    \toprule
        & \multicolumn{3}{c}{Hyperopt} & \multicolumn{3}{c}{Guidance DI} & \multicolumn{3}{c}{Guidance F1} \\
        \cmidrule(rl){2-4} \cmidrule(rl){5-7} \cmidrule(rl){8-10}
    Dataset         & Pipeline  & F1 & DI & Pipeline  & F1 & DI & Pipeline  & F1 & DI \\
    \midrule
    COMPAS V. & Pr(Stack(e, e))           & 0.25 \stdevfont(0.06) & 0.86 \stdevfont(0.13) & In            & \closetohopt{0.35} \stdevfont(0.07) & \closetohopt{0.76} \stdevfont(0.10) & Bag(In, 10)     & 0.0  \stdevfont(0.0)  & 0.0  \stdevfont(0.0) \\
    Credit-g       & Vote(Pr(e))          & 0.85 \stdevfont(0.01) & 0.99 \stdevfont(0.04) & In            & \vclosetohopt{0.83} \stdevfont(0.01) & \vclosetohopt{0.96} \stdevfont(0.16) & Bag(In, 10)     & \vclosetohopt{0.84} \stdevfont(0.01) & \vclosetohopt{0.93} \stdevfont(0.09) \\
    COMPAS          & Bag(Post(e), 72)    & 0.56 \stdevfont(0.02) & 0.94 \stdevfont(0.01) & In            & 0.18 \stdevfont(0.28) & \closetohopt{0.88} \stdevfont(0.20) & Bag(In, 10)     & 0.0  \stdevfont(0.0)  & 0.0  \stdevfont(0.0) \\
    Ricci          & Pr(e)                & 0.82 \stdevfont(0.09) & 1.06 \stdevfont(0.20) & In            & \vclosetohopt{0.80} \stdevfont(0.07) & 0.76 \stdevfont(0.28) & Bag(In, 10)     & \vclosetohopt{0.86} \stdevfont(0.04) & 0.58 \stdevfont(0.16) \\
    TAE             & In                  & 0.34 \stdevfont(0.29) & 1.00 \stdevfont(0.00) & Post(e)       & 0.58 \stdevfont(0.08) & 0.54 \stdevfont(0.16) & Boost(Pr, 500)  & 0.71 \stdevfont(0.06) & 0.51 \stdevfont(0.23) \\
    Titanic         & Pr(Vote(e))         & 0.73 \stdevfont(0.06) & 0.74 \stdevfont(0.22) & Post(e)       & 0.87 \stdevfont(0.02) & 0.36 \stdevfont(0.02) & Boost(Pr, 500)  & 0.89 \stdevfont(0.04) & 0.38 \stdevfont(0.05)\\
    SpeedDating     & In                  & 0.44 \stdevfont(0.01) & 1.02 \stdevfont(0.28) & In            & \vclosetohopt{0.48} \stdevfont(0.03) & \closetohopt{0.94} \stdevfont(0.14) & Pr(Stack(e, e)) & 1.0  \stdevfont(0.0)  & 0.85 \stdevfont(0.02)\\
    Bank            & Post(Boost(e, 206)) & 0.94 \stdevfont(0.00) & 1.00 \stdevfont(0.00) & In            & \vclosetohopt{0.95} \stdevfont(0.00) & \vclosetohopt{0.99} \stdevfont(0.01) & Pr(Stack(e, e)) & \closetohopt{0.84} \stdevfont(0.01) & 0.74 \stdevfont(0.04) \\
    MEPS 19        & In                   & 0.30 \stdevfont(0.01) & 0.83 \stdevfont(0.03) & In            & \vclosetohopt{0.26} \stdevfont(0.20) & 1.02 \stdevfont(0.21) & Pr(Stack(e, e)) & \vclosetohopt{0.25} \stdevfont(0.15) & 0.70 \stdevfont(0.10)\\
    MEPS 20        & Stack(e, In)         & 0.32 \stdevfont(0.01) & 0.80 \stdevfont(0.06) & Boost(In, 50) & 0.43 \stdevfont(0.02) & \vclosetohopt{0.78} \stdevfont(0.02) & Boost(In, 10)   & 0.43 \stdevfont(0.01) & \vclosetohopt{0.78} \stdevfont(0.05) \\
    Nursery         & Pr(e)               & 0.74 \stdevfont(0.12) & 1.03 \stdevfont(0.10) & Boost(In, 50) & \vclosetohopt{0.78} \stdevfont(0.01) & 0.85 \stdevfont(0.03) & Boost(In, 10)   & \vclosetohopt{0.78} \stdevfont(0.02) & 0.86 \stdevfont(0.10) \\
    MEPS 21        & In                   & 0.38 \stdevfont(0.09) & 0.86 \stdevfont(0.02) & Boost(In, 50) & \vclosetohopt{0.43} \stdevfont(0.01) & \closetohopt{0.76} \stdevfont(0.07) & Boost(In, 10)   & \vclosetohopt{0.43} \stdevfont(0.01) & 0.75 \stdevfont(0.07)\\
    Adult           & Pr(e)               & 0.66 \stdevfont(0.01) & 0.29 \stdevfont(0.06) & Boost(In, 50) & 0.09 \stdevfont(0.01) & 0.81 \stdevfont(0.15) & Boost(In, 10)   & 0.09 \stdevfont(0.01) & 0.81 \stdevfont(0.15)\\
    \bottomrule
    \end{tabular}
    \normalsize\par
    \caption{Recommended configurations for each dataset from Hyperopt search, guidance diagram 
    to optimize for fairness, and guidance diagram to optimize for predictive performance,
    along with average and standard deviation (in parentheses) of F1 and DI.
    Average guidance diagram values in \vclosetohopt{green} are within 0.05 of the corresponding average Hyperopt value for the same metric and dataset,
    and average values in \closetohopt{blue} are within 0.1 of the corresponding average Hyperopt value. 
    }
    \label{tab:hyperopt-and-guidance-recs}
\end{table*}

\normalsize\par

%% file: fig_tree.tex
\begin{figure*}[!t]
\centerline{\includegraphics[width=\textwidth]{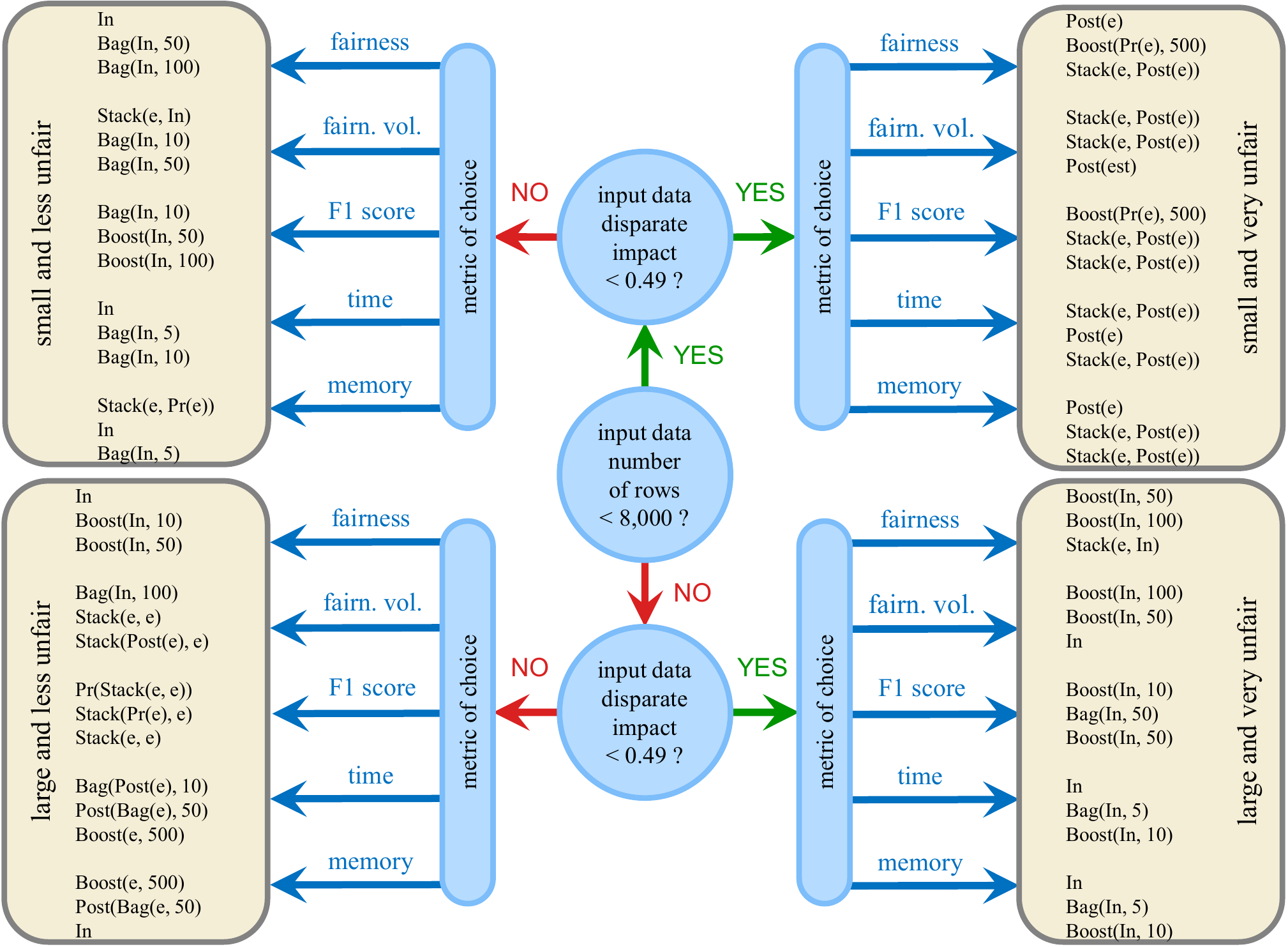}}
\caption{Guidance diagram for picking a good starting configuration given dataset characteristics and a target metric.}
\label{fig:tree}
\end{figure*}

%% file: discussion.tex
\section{Discussion}

This section describes the impacts of our search results and guidance
in addition to hypotheses informed by our results regarding biased data.

\paragraph{Guidance diagram utility and robustness.}

The previous section showed that the guidance diagram and Hyperopt search
recommended configurations with relatively similar performance on most
of the datasets. This suggests that the guidance diagram can be useful
for future practitioners to recommend starting points for model development
based on their data setup and metric(s) of interest. Moreover, note that 
consulting the guidance diagram for a configuration that may likely produce
reasonable results can be done quickly without any resource consumption,
while searching may require
inordinate amounts of time and compute. 

Our leave-one-out dataset experiments also suggest
that our diagram generation algorithm is relatively robust to changes
in the input data. This further supports the notion that our guidance
diagram has useful recommendations. However, those experiments also
showed that the presence or absence of certain datasets affected
the resulting diagram more than others. For instance, the Credit-g and Bank
datasets have more effects on the recommended configurations and model
performance than the Adult or COMPAS datasets.

We attribute this phenomenon to the filtering of model results that takes
place during diagram generation and properties of the datasets themselves.
Note that most of the datasets in Table \ref{tab:dataset-table}
that have large effects
on the diagram have baseline disparate impact close to 0.8 (meaning they 
are relatively fair), and their protected attributes are not strongly
predictive (based on feature importance ranking). This implies
that with mitigation, it is possible to fit these datasets fairly
and accurately. This in turn means model fitting results from those datasets
comprise most of the results for the given quadrant after filtering
to reasonable fairness and predictive performance. Therefore,
when those datasets are missing, the generated diagram differs greatly
from the one generated with all data. (The exceptions
to this rule are TAE and Titanic. Given that those are the only
two datasets in their quadrant and protected attributes are strongly
predictive, it is difficult to fit either well in a fair manner.
Therefore, neither contributes many fitting results after filtering,
and both have tangible effects on the diagram.)

In light of how protected attribute feature importance
of input datasets affects recommendations of the guidance diagram,
one limitation of our diagram is that we do not have branches
for this property of a dataset (and therefore do not provide
recommendations based on this property). We counter that
determining this property requires training
XGBoost models, which can take time and resources, while 
the other properties utilized can be quickly calculated.
Thus, we still argue
that this is useful to future practitioners.

\paragraph{What is ``good data?''}

As mentioned in \citet{holstein_et_al_2019},
``future research'' in the area of algorithmic fairness should
``[develop] processes and tools for fairness-focused debugging'' and
``should also support practitioners in collecting and curating 
high-quality datasets in the first place''. These recommendations suggest
\emph{how to collect good data?} and \emph{what even \emph{is} `good data'?}
are questions with which the field is currently grappling.

We believe our results shed some light on these fronts. First, our findings
suggest that converting ``bad data'' to ``good data'' may not (just) involve
making datasets larger in number of \emph{examples} but (also) 
making them larger in
number of \emph{features}. Prevailing notions of algorithmic fairness
may imply that 
the best way to fix an unfair dataset is to add examples to
reduce bias. While this may work, it could be difficult
to do in practice (especially given societal mechanisms behind bias),
and \citet{holstein_et_al_2019} also raise that ``How much data [one] would
need to collect?'' does not typically have a clear answer. However,
our results imply that collecting more data
to alleviate bias may take the form of gathering more features instead of 
simply gathering more examples with the same features.

That being said, datasets like Adult and Ricci included attributes that were
more predictive than protected attributes, yet they still did not strongly
influence our guidance diagram. We conjecture that attributes that were
more predictive were highly correlated with protected attributes, and feature
importance tables included in our Appendix seem to support this notion.
Therefore, if one wants to collect more features to reduce bias, they should
take care to ensure that these features are not correlated with protected attributes.

Lastly, we want to highlight that regardless of the form such data collection
may take, it is imperative to consider the ethics of doing so and 
respect wishes and privacy of the individuals whose data are utilized during the 
model building process and who are affected by the model predictions.

%% file: conclusion.tex
\section{Conclusion}\label{sec:conclusion}

This paper introduces a library of modular bias mitigators and ensembles
and detail experiments that confirm ensembles can improve fairness stability.
We also provide generalizable guidance to practitioners based on their data setup.

%% file: appendix.tex
\appendix
\section{Supplemental Material}\label{sec:supplemental}

\subsection{Additional Tables}
\begin{table}[!h]
    \begin{tabular}{ll}
    \toprule
    Dataset         & Protected Attribute Rankings \\
    \midrule
    COMPAS Violent   & \makecell{sex: 4 \\ race: 7}                         \\
    Credit-g        & \makecell{age: 22 \\ sex: 43}                          \\
    COMPAS            & \makecell{sex: 5 \\ race: 7}                          \\
    Ricci            & \makecell{race: 6 \\}                                  \\
    TAE               & \makecell{native\_english\_speaker: 1 \\}              \\
    Titanic           & \makecell{sex: 2 \\}                                  \\
    SpeedDating      & \makecell{importance\_same\_race: 24 \\ samerace: 69}  \\
    Bank             & \makecell{age: 17 \\}                                   \\
    MEPS 19         & \makecell{RACE: 22 \\}                                  \\
    MEPS 20         & \makecell{RACE: 18 \\}                                  \\
    Nursery           & \makecell{parents: 3 \\}                              \\
    MEPS 21         & \makecell{RACE: 10 \\}                                 \\
    Adult            & \makecell{sex: 19 \\ race: 30}                        \\
    \bottomrule
    \end{tabular}
    \caption{Granular feature importance rankings of protected attributes for each dataset.}
    \label{tab:protected-attr-feature-imp}
\end{table}

\begin{table}[!h]
    \setlength{\tabcolsep}{2.6pt}
    \begin{tabular}{lll}
    \toprule
    & \multicolumn{2}{c}{Feature 1}\\
    \cmidrule(rl){2-3}  
    Dataset         & Name  & Imp.  \\
    \midrule
    COMPAS V. & priors\_count=More than 3          & 0.49              \\
    Credit-g & checking\_status\_no checking      & 0.11             \\
    COMPAS & priors\_count=More than 3          & 0.54              \\
    Ricci & combine                            & 1                        \\
    TAE & native\_english\_speaker           & 0.33               \\
    Titanic & boat\_13                           & 0.79  \\
    SpeedDating & like                               & 0.08 \\
    Bank & poutcome\_success                  & 0.18              \\
    MEPS 19 & WLKLIM=2.0                         & 0.34          \\
    MEPS 20 & WLKLIM=2.0                         & 0.26             \\
    Nursery & health\_not\_recom                 & 0.59            \\
    MEPS 21 & WLKLIM=2.0                         & 0.19          \\
    Adult & marital-status\_Married-civ-spouse & 0.42              \\
    \bottomrule
    \end{tabular}
    \caption{Feature importance information for first-ranked feature for each dataset.}
\end{table}

\begin{table}[!h]
    \setlength{\tabcolsep}{2.6pt}
    \begin{tabular}{lll}
    \toprule
    & \multicolumn{2}{c}{Feature 2}\\
    \cmidrule(rl){2-3}  
    Dataset         & Name  & Imp.  \\
    \midrule
    COMPAS V.     & age\_cat=Less than 25            & 0.22                   \\
    Credit-g         & other\_parties\_guarantor        & 0.03                   \\
    COMPAS             & age\_cat=Less than 25            & 0.25                   \\
    Ricci            & position\_Captain                & 0                          \\
    TAE              & summer\_or\_regular\_semester\_1 & 0.29                   \\
    Titanic          & sex                              & 0.04                   \\
    SpeedDating      & attractive\_o                    & 0.05                   \\
    Bank              & contact\_unknown                 & 0.08                    \\
    MEPS 19          & ARTHDX=1.0                       & 0.06                   \\
    MEPS 20          & ARTHDX=1.0                       & 0.06                   \\
    Nursery          & has\_nurs\_very\_crit            & 0.08                   \\
    MEPS 21          & ARTHDX=1.0                       & 0.10                   \\
    Adult            & education-num                    & 0.05                  \\
    \bottomrule
    \end{tabular}
    \caption{Feature importance information for second-ranked feature for each dataset.}
\end{table}

\begin{table}[!h]
    \setlength{\tabcolsep}{2.6pt}
    \begin{tabular}{lll}
    \toprule
    & \multicolumn{2}{c}{Feature 3}\\
    \cmidrule(rl){2-3}  
    Dataset         & Name  & Imp. \\
    \midrule
    COMPAS V.      & age\_cat=Greater than 45  & 0.15      \\
    Credit-g           & credit\_history\_all paid & 0.03     \\
    COMPAS               & age\_cat=Greater than 45  & 0.07      \\
    Ricci                & position\_Lieutenant      & 0            \\
    TAE                   & course                    & 0.15     \\
    Titanic               & boat\_A                   & 0.04     \\
    SpeedDating          & funny\_o                  & 0.04     \\
    Bank                 & month\_mar                & 0.05     \\
    MEPS 19             & ACTLIM=1.0                & 0.03     \\
    MEPS 20             & INSCOV=3.0                & 0.02     \\
    Nursery               & parents                   & 0.07     \\
    MEPS 21             & ACTLIM=2.0                & 0.04      \\
    Adult                & capital-gain              & 0.05    \\
    \bottomrule
    \end{tabular}
    \caption{Feature importance information for third-ranked feature for each dataset.}
\end{table}

\begin{table}[!h]
    \setlength{\tabcolsep}{2.6pt}
    \begin{tabular}{lll}
    \toprule
    & \multicolumn{2}{c}{Feature 4}\\
    \cmidrule(rl){2-3}  
    Dataset         & Name  & Imp. \\
    \midrule
    COMPAS V.      & sex                               & 0.05                \\
    Credit-g           & savings\_status\_no known savings & 0.03                \\
    COMPAS               & priors\_count=0                   & 0.06                \\
    Ricci               & oral                              & 0                       \\
    TAE                  & course\_instructor                & 0.13                \\
    Titanic              & parch                             & 0.02                \\
    SpeedDating       & attractive\_partner               & 0.03                \\
    Bank                & month\_jun                        & 0.04                 \\
    MEPS 19            & ADSMOK42=-1.0                     & 0.02                \\
    MEPS 20            & ACTLIM=1.0                        & 0.02                \\
    Nursery              & has\_nurs\_critical               & 0.06                \\
    MEPS 21            & ACTLIM=1.0                        & 0.03                \\
    Adult               & occupation\_Other-service         & 0.03               \\
    \bottomrule
    \end{tabular}
    \caption{Feature importance information for fourth-ranked feature for each dataset.}
\end{table}

\begin{table}[!h]
    \setlength{\tabcolsep}{1pt}
    \begin{tabular}{lll}
    \toprule
    & \multicolumn{2}{c}{Feature 5}\\
    \cmidrule(rl){2-3}  
    Dataset         & Name  & Imp. \\
    \midrule
    COMPAS V.    & age\_cat=25 to 45                      & 0.03                  \\
    Credit-g         & property\_magnitude\_no known property & 0.03                  \\
    COMPAS             & sex                                    & 0.03                  \\
    Ricci             & written                                & 0                         \\
    TAE                & class\_size                            & 0.11                  \\
    Titanic            & body                                   & 0.02                  \\
    SpeedDating      & funny\_partner                         & 0.03                  \\
    Bank              & duration                               & 0.04                  \\
    MEPS 19          & JTPAIN=1.0                             & 0.02                  \\
    MEPS 20          & ACTLIM=2.0                             & 0.02                  \\
    Nursery            & has\_nurs\_improper                    & 0.03                  \\
    MEPS 21          & INSCOV=3.0                             & 0.02                  \\
    Adult             & relationship\_Own-child                & 0.03                 \\
    \bottomrule
    \end{tabular}
    \caption{Feature importance information for fifth-ranked feature for each dataset.}
\end{table}

\input{hyperparam_tables}

%% file: hyperparam_tables.tex
\begin{table*}[!h]
    \begin{tabular}{lll}
    \toprule
    Dataset        & Mitigator              & Hyperparameters            \\
    \midrule
    COMPAS Violent & DisparateImpactRemover & 1                          \\
    Credit-g       & LFR                    & k=5, Ax=0.01, Ay=10, Az=5  \\
    COMPAS         & DisparateImpactRemover & 0.4                        \\
    Ricci          & LFR                    & k=5, Ax=0.01, Ay=5, Az=10  \\
    TAE            & LFR                    & k=5, Ax=0.01, Ay=50, Az=5  \\
    Titanic        & DisparateImpactRemover & 0.8                        \\
    SpeedDating    & DisparateImpactRemover & 0.2                        \\
    Bank           & DisparateImpactRemover & 0.2                        \\
    MEPS 19        & LFR                    & k=5, Ax0.01, Ay=1, Az=10   \\
    MEPS 20        & LFR                    & k=5, Ax=0.01, Ay=1, Az=10  \\
    Nursery        & LFR                    & k=20, Ax=0.01, Ay=1, Az=10 \\
    MEPS 21        & LFR                    & k=5, Ax=0.01, Ay=1, Az=10  \\
    Adult          & LFR                    & k=5, Ax=0.01, Ay=1, Az=10  \\
    \bottomrule
    \end{tabular}
    \caption{Optimal pre-estimator mitigator configurations (with corresponding hyperparameters) per dataset.
    Hyperparameter names are not provided if the mitigation technique only accepts one.
    If a hyperparameter is not listed in the rightmost column, the configuration utilizes the default value.}
    \label{tab:pre-est-configs}
\end{table*}

\begin{table*}[!h]
    \begin{tabular}{lll}
    \toprule
    Dataset        & Mitigator            & Hyperparameters                    \\
    \midrule
    COMPAS Violent & MetaFairClassifier   & 0.5                                \\
    Credit-g       & AdversarialDebiasing & classifier\_num\_hidden\_units=10  \\
    COMPAS         & MetaFairClassifier   & 0.5                                \\
    Ricci          & MetaFairClassifier   & 0.8                                \\
    TAE            & MetaFairClassifier   & 0.8                                \\
    Titanic        & MetaFairClassifier   & 1                                  \\
    SpeedDating    & MetaFairClassifier   & 0.9                                \\
    Bank           & PrejudiceRemover     & 100                                \\
    MEPS 19        & PrejudiceRemover     & 1000                               \\
    MEPS 20        & AdversarialDebiasing & classifier\_num\_hidden\_units=500 \\
    Nursery        & MetaFairClassifier   & 0.5                                \\
    MEPS 21        & AdversarialDebiasing & classifier\_num\_hidden\_units=500 \\
    Adult          & PrejudiceRemover     & 1000                               \\
    \bottomrule
    \end{tabular}
    \caption{Optimal in-estimator mitigator configurations (with corresponding hyperparameters) per dataset.
    Hyperparameter names are not provided if the mitigation technique only accepts one.
    If a hyperparameter is not listed in the rightmost column, the configuration utilizes the default value.}
    \label{tab:in-est-configs}
\end{table*}